\documentclass[preprint,12pt]{elsarticle}

\usepackage{hyperref}
\usepackage{amsmath,amssymb,mathtools,float,times,bm,proof,tabularx,capt-of,natbib}
\usepackage{booktabs}
\usepackage{subcaption}
\usepackage{adjustbox}
\usepackage{comment}
\usepackage{multirow}
\usepackage{xcolor}
\usepackage{wrapfig}

\journal{Journal of Biomedical Informatics}

\begin{document}

\begin{frontmatter}



\title{Neural Network Architecture for Database Augmentation Using Shared Features}



\author[ro]{William C. Sleeman IV}
\author[ro]{Rishabh Kapoor}
\author[cs]{Preetam Ghosh}

\address[ro]{Virginia Commonwealth University, Department of Radiation Oncology, Richmond, VA }
\address[cs]{Virginia Commonwealth University, Department of Computer Science, Richmond, VA}

\begin{abstract}
The popularity of learning from data with machine learning and neural networks has lead to the creation of many new datasets for almost every problem domain. However, even within a single domain, these datasets are often collected with disparate features, sampled from different sub-populations, and recorded at different time points. Even with the plethora of individual datasets, large data science projects can be difficult as it is often not trivial to merge these smaller datasets. Inherent challenges in some domains such as medicine also makes it very difficult to create large single source datasets or multi-source datasets with identical features. Instead of trying to merge these non-matching datasets directly, we propose a neural network architecture that can provide data augmentation using features common between these datasets. Our results show that this style of data augmentation can work for both image and tabular data.
\end{abstract}


\begin{keyword}


neural networks \sep machine learning \sep databases \sep autoencoders \sep classification

\end{keyword}

\end{frontmatter}


\section{Introduction}
\label{sec:introduction}
Data analysis tools from machine learning and artificial intelligence are now used to solve problems within almost every industry and problem domain. Although more data is continuously being collected to support those advanced methods, real world data science problems are often challenged with a limited number of examples or missing features. For example, collecting medical data is costly and time consuming as it requires patient consent, data security for protecting privacy, and the need for subject matter experts. If these challenges are addressed, collecting large patient cohorts still may not be possible as inclusion for a given study may be restrictive or the medical center may not treat enough such patients each year. Even outside of medicine, many of the available datasets are still relatively small. Within the popular UC Irvine Machine Learning Repository (UCI), half of the datasets contain fewer than 1,600 examples \cite{Dua:2019}.

To make the most of modern machine learning and deep learning algorithms, quality data is needed for training and the more data available often produces the best results. Very large, data-centric companies have the capability to construct massive datasets but this is often not scalable to many problem domains. Collecting, cleaning, and storing large datasets is expensive and smaller companies or medical studies focusing on a single diagnosis simply may have no way to generate large amounts of data. Aggregating across multiple data sources is one solution to address small datasets but becomes non-trivial when features do not match, some of the data is missing, or different feature encoding is used. However, it is likely some of the features are present for multiple datasets if they belong to the same problem domain. Medical data often includes demographics information like age, sex, race, or diagnosis and other domains may use industry standards resulting in feature overlap.

In addition to the common features, each dataset likely has other unique features that make it difficult for direct aggregation. However, the shared context between these datasets may provide enough information to transfer the unique context between these non-matching datasets. Our approach uses autoencoder networks to convert common features from a given dataset (\textbf{A}) to its full complement of features. The same common features from a different dataset (\textbf{B}) can then be passed through \textbf{A}'s network to generate synthetic features for dataset \textbf{A}. The unique features from the new synthetic examples can then be added to the existing features for database \textbf{B}, creating a new augmented dataset.

The primary motivation for this work is to address the challenges faced when trying to learn from relatively small medical datasets. As previously mentioned, it is often impracticable to create large datasets focusing on specific medical questions. There are many high quality medical datasets and we hypothesized that predictive performance would be improved by including information from other datasets based on the common features. In Section \ref{sec:gdc-seer-results}, we discuss how this data augmentation method can be applied to real world medical datasets.

In summary, we propose an autoencoder based approach for augmenting datasets using common features. Our experiments include both CNN networks for images and fully connected networks for tabular data, providing insight on how feature sharing impacts performance. We also show that this method can improve classifier performance, specifically with the use case of cancer datasets.

This work addresses the challenge of combining knowledge across disparate datasets and our main contributions are the following:
\begin{itemize}
	\item \textbf{Proposed architecture:} We propose an autoencoder based solution for augmenting dataset using common features. This provides a method to create synthetic examples that are completely missing from a dataset which adds more context.
	\item \textbf{Study of the impact of feature sharing:} This initial experimental study is performed on both image and tabular based datasets and the impact of data sharing is investigated.
	\item \textbf{Use case with clinical datasets:} A real world example of cancer data is used to show the proposed method's performance. 
	\item \textbf{Software:} We provide a publicly available software package using Python with Keras on GitHub for future research.
\end{itemize}

\section{Related Works}
\label{sec:related_works}
One challenge with real world machine learning and deep learning projects is the limited size and   quality of training data. Some domains like medicine are notorious for the difficulty of building large datasets with concerns of privacy, data collection cost, and regulatory restrictions. Even when enough data is present, issues like class imbalance can reduce model performance. 

Data augmentation is often used to increase the size of a dataset or improve data quality. One of the most common ways to augment image data is to apply shifts, rotations, color adjustment, or zooming. This keeps the main concepts of the image but introduces enough variation to help with overfitting while providing an almost unlimited number of permutations. More recent advances in image augmentation include patch cutout, blurring, and image mixing where two training examples are blended using various methods \cite{lewy2022overview, jadhav2023study}. The XtremeAugment method was also developed to generate new examples by adding one or more training objects with adjusted color or viewpoint, all on existing or new backgrounds \cite{nesteruk2022xtremeaugment}. 

Instead of perturbing existing data, completely new synthetic examples can also be generated. Techniques like random oversampling, Synthetic Minority Oversampling Technique (SMOTE) \cite{chawla2002smote}, including its many derived methods, can be used to add more examples to the training data. Although those algorithms were originally designed for traditional machine learning problems, DeepSMOTE \cite{dablain2022deepsmote} was later created for deep learning problems. Unlike traditional data augmentation and random oversampling, some of the SMOTE based methods can create examples in specific portions of the feature space. This allows for placing more focus on areas of interest like decision boundaries, safe regions, or regions specifed by clustering algorithms \cite{han2005borderline, bunkhumpornpat2009safe, douzas2018improving}.  Undersampling can also be used for class balancing \cite{lin2017clustering, arefeen2020neural} by reducing the size of the majority class. Both over- and undersampling can be done within the same algorithm shown by the Combined Synthetic Oversampling and Undersampling Technique for Imbalanced Data Classification (CSMOUTE) method \cite{koziarski2021csmoute}.

Another solution for creating larger datasets is to combine multiple smaller dataset that include the same kind of information. However, these datasets may not share the same features or have the same data ranges. Several works proposed solutions to database merging including a method that combined multiple image datasets using only their principal component analysis (PCA) space and did not require the original training images to be kept \cite{costache2009combining}. PCA was used for feature reduction which could aid in combining datasets with many non-shared features \cite{nguyen2022combining}. Another work showed that singular value decomposition (SVD) could be used to combine partially overlapping microarray gene expression datasets \cite{schreiber2008combining}. 

The occurrence of missing feature values is a common challenge as one survey suggested that over half of the UCI datasets had a missing feature rate of at least 30\% \cite{lin2020missing}. Imputation is often used to replace missing values which can be based on methods like simple statistics, regression, hot-deck method, clustering, and other machine learning algorithms \cite{emmanuel2021survey}. Neural networks have been used for imputation with both traditional autoencoders \cite{choudhury2019imputation, beaulieu2017missing} and GANs \cite{shang2017vigan}.

\section{Proposed Architecture}
\label{sec:arch}

Applying traditional data augmentation methods may result in sub-optimal results when faced with disparate datasets. Model generalization and class imbalance can be partially addressed with synthetic examples generated from a single source but this excludes any novel information present in other relevant datasets. Feature reduction methods may remove critical relationships within smaller classes or interesting regions of the feature space. To address some of these limitations, we propose a neural network based method that generates unseen features from common ones, thereby improving the predictive power of learning algorithms.

\begin{figure}
	\centering
	\includegraphics[width=1.0\linewidth,trim=5 5 5 5,clip]{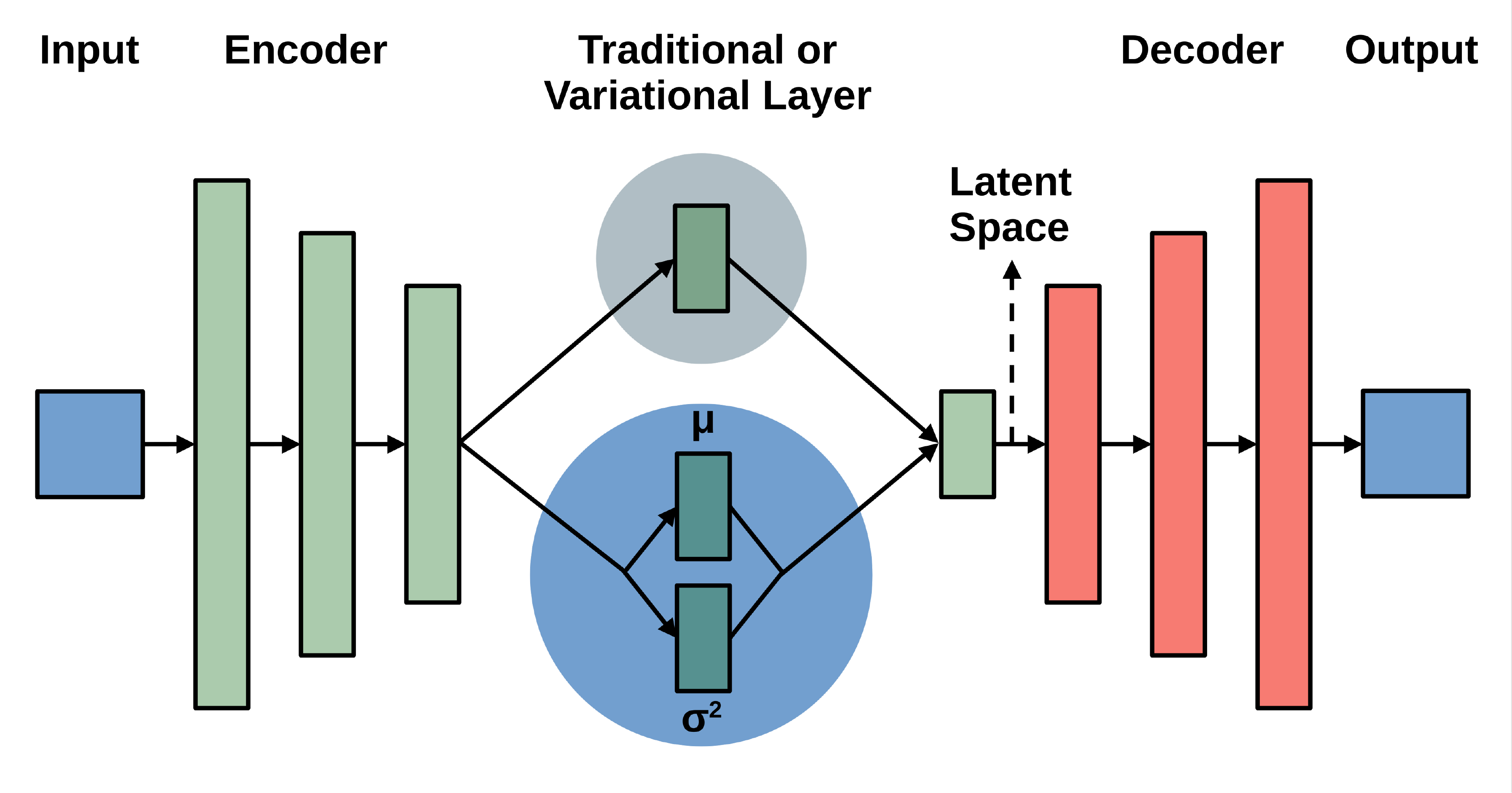}
	\caption{An example autoencoder network, broken down into the input, encoder, decoder, and output sections. This architecture varies between the methods used in the experimental study with AE shown in gray and VAE in blue.}
	\label{fig:basic_ae}
\end{figure}

The core component is an autoencoder network and we investigated the use of both a traditional autoencoder (AE) and a variational autoencoder (VAE) architecture. Figure \ref{fig:basic_ae} shows the general architecture used in the experiments found in Section \ref{sec:experiments}. Input data is passed through encoding layer and compressed into the latent space. The primary difference between these two architectures is the step right before the latent space layer as depicted in gray for AE and blue for VAE. While the AE uses a layer like the prior encoder layers, the VAE splits into mean ($\mu$) and variance ($\sigma^{2}$) sub-layers. A stochastic sampling method is then used to produce the layer output when generating the latent values. This also means that the same input data may result in different outputs after the model is frozen unlike the traditional AE.

If two datasets, referred to \textit{A} and \textit{B}, represent different aspects of the same concept, the unique features from each dataset may be correlated. This architecture uses the common features to generate synthetic replacements for feature values that are only present in the other dataset. By combining the existing and synthetic features, the new examples simulate the scenario where all features came from a single complete data source. In the following description of the processing steps, we are augmenting dataset \textit{A} with the unique features in dataset \textit{B}. Below are the detailed steps of this process:\newline

\noindent\textbf{Identify Common Features:}
First, the features common between the two datasets are identified. Min-max normalization is performed on the common features across both datasets to ensure proper alignment and the same process is used for the dataset specific columns resulting in a range of [0, 1]. Next, new sub-datasets named \textit{CA} and \textit{CB} are extracted from the pre-processed \textit{A} and \textit{B}, representing the common features of the examples present in each dataset.\newline

\noindent\textbf{Fit the Autoencoder Network with \textit{B}:}
An autoencoder network is trained to map \textit{CB} data to the full compliment of features found in \textit{B}.\newline

\noindent\textbf{Predict with \textit{B} Data:}
The autoencoder is frozen and the \textit{CA} is passed through the \textit{CB-B} network to predict the \textit{B} features values.
\newline

\noindent\textbf{Append Generated \textit{B} Features:}
Database \textit{A} is now augmented with the \textit{B} synthetic features, adding information from the similar, but independent, database \textit{B}.\newline

\begin{figure}
	\centering
	\includegraphics[width=0.5\linewidth,trim= 5 5 5 5,clip]{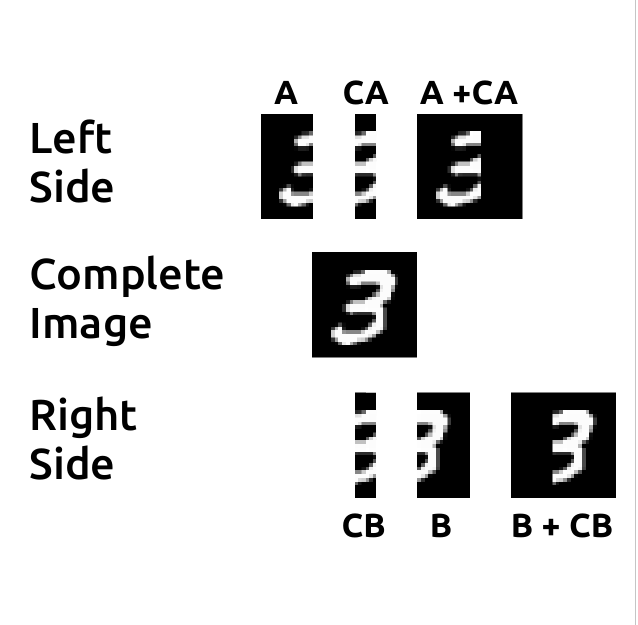}
	\caption{Example of how images could be split to simulate two different datasets. Right hand (\textit{A}) and its associated middle common columns (\textit{CA}) go to the masked database \textit{A} image and the left hand (\textit{B}) with its common columns (\textit{CB}) go to database \textit{B}.}
	\label{fig:imagesplitting}
\end{figure}

To better visualize this process, we show in Figure\ref{fig:imagesplitting} how the MNIST image data was split by image columns to simulate the two dataset problem considered above. Dataset \textit{A} is given the left half and dataset \textit{B} is given the right half of the MNIST image columns. 

Common features between these new datasets are simulated by reintroducing some columns from the opposing dataset. To make the experiments more consistent, the resulting training images are padded with zero columns so the final dimensions are the same as in the original dataset. The example in Figure \ref{fig:imagesplitting} shows this process with six common columns, with first row showing the dataset \textit{A} with the 14 left hand columns, the six common columns, and the result of combining left hand and common columns with padding. The process is performed for the right hand dataset \textit{B} with a comparison to the original image in the middle.

\begin{figure}
	\centering
	\includegraphics[width=0.8\linewidth,trim= 2 3 2 3,clip]{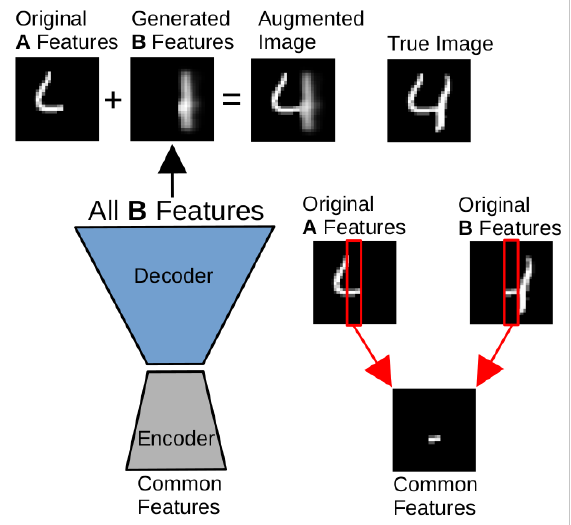}
	\caption{The proposed data augmentation network. Starting at the bottom, common columns of the masked training example from dataset \textit{B} are used to train an autoencoder network. Test data from dataset \textit{A} is processed using the \textit{CB-B} network to get synthetic B features. These new features are added to the \textit{A} masked test data to generate the augmented image.}
	\label{fig:architecture}
\end{figure}

Figure \ref{fig:architecture} shows an entire diagram of the architecture. Using the split image data described above, the common columns are used to train on the \textit{CB-B} network. Common columns from the test \textit{A} dataset are encoded using the \textit{CB-B} network and decoded into synthetic \textit{B} features. The original \textit{A} data is then combined with the synthetic features to create the augmented image. While the resulting image is not perfect, it gets closer to replicating the original image.

\newpage

\section{Experimental Study}
\label{sec:experiments}
In this section, we apply the data augmentation process to both image and tabular data using traditional and variational autoencoder networks. Source code used for these experiments is provided at https://github.com/fsleeman/database-augmentation-network which was written is Python with Keras.

\subsection{MNIST and CIFAR10}
\label{sec:imagingData}
In these first experiments, we test the augmentation process using the MNIST \cite{lecun1998mnist} and CIFAR10 \cite{krizhevsky2009learning} datasets. The image features were split using the same process as shown in Figure \ref{fig:imagesplitting} to simulate datasets with unavailable data. For the purpose of these experiments, each image column was treated as if it was as a single feature. The middle columns that overlap with the left and right sides of the original images were marked as common features (shown in Figure \ref{fig:imagesplitting} as CA and CB). An even number of common columns (\textit{n}) were used to provided a symmetry that allowed for the same experiments to be performed on both sides of the images. We performed experiments between 2 and \textit{total columns} – 2 for both datasets, where MNIST had 28 total columns and CIFAR10 had 32. Future experiments can investigate other combinations of feature sharing but the small size of MNIST and CIFAR10 images limits the number of useful combinations. 

The impact of the data augmentation process was evaluated using a simple CNN based classifier. This model used two CNN layers with max pooling, flattened, a with dropout, a dense layer, one more dropout layer, and  a final softmax layer. Both MNIST and CIFAR10 datasets had ten classes and so argmax was used to choose the predicted class.	The only difference between how these datasets were processed was that the MNIST networks used one color channel as included gray scale images and the CIFAR10 used three color channels for its RGB color. Twenty percent of the data was held out for testing and the remaining data was used for training and validation. To better represent two completely independent datasets, the training data was split, so database \textit{A} got the first half of the examples (by index) and database \textit{B} got the second half. This meant that there was no information overlap between those datasets including the common features. Training was then performed using 5-fold cross validation and the remaining test data was used to provide the results in Sections \ref{sec:results:mnist} and \ref{sec:results:cifar10}.

\subsection{Lung Cancer Case Study}
\label{sec:experiments:nihseer}
One of the major problems faced with clinically oriented data science projects is the lack of data. While there are many high quality datasets in the field of medicine, they often do not have the same features or formatting that would allow them to be combined directly. One motivation of this work was to devise an approach that would improve model performance with information from independent medical datasets. In these experiments, we apply the proposed data augmentation method on two disjointed lung cancer datasets.

The National Institutes of Health (NIH) has provided a number of cancer related datasets including the Genomic Data Commons (GDC) \cite{grossman2016toward} and the Surveillance, Epidemiology, and End Results (SEER) Program \cite{SEER}. Started in 2016, the GDC is a harmonized cancer dataset which combines data from several modalities such as gene expression, mutations, pathology images, prescribed drugs, and clinical outcomes across over eighty six thousand patients. SEER has been collecting cancer data since 1973 from cancer registries across the United States which currently has over fifteen million reported cases.

We have chosen to limit the data in our experiments to lung cancer because it is one of the most common diagnoses and has a wide range of survival outcomes. Unlike the SEER dataset which is in a tabular format, GDC is mostly made up of other modalities which further reduced the amount of data used. After filtering for lung cancer, samples with missing values were then removed. There are many methods for addressing missing features but these examples were removed because we wanted to experiment on the cleanest available data. Other approaches in handling missing features should be further investigated.

The data used in the following experiments resulted in 522 examples for GDC and 674,008 for SEER after the filtering and data cleaning. Seven common features between the two cleaned datasets were then identified as shown in Table \ref{table:common_columns}: sex, year of diagnosis, age at diagnosis, race, International Classification of Diseases version 10 (ICD-10) code, histology, and laterality. Categorical features were split into multiple columns using one-hot encoding and min-max scaling was performed across both datasets for normalization. After one-hot encoding, there were a total of 27 individual common features. 

The unique features of the two datasets covered other relevant lung cancer information that likely would add value to the learning process. GDC included details such as smoking history, pathological staging, and ethnicity while SEER included group staging, income, rural vs. urban locale, surgery type, the number of tumors, among others. These unique features were also min-max normalized, resulting in a total of 68 features for GDC and 78 for SEER. 

\begin{table}[h!]
	\centering
	\renewcommand{\arraystretch}{1.25}
	\begin{adjustbox}{center,max width = 100mm}
		\footnotesize
		\begin{tabular}{*{3}{l}}\hline
			\multicolumn{3}{c}{SEER and NIH Common Features}\\
			\bottomrule
			Feature & Type & Possible Values \\
			\hline
			Sex & Categorical & Female, Male \\
			Year of diagnosis & Numerical & 1957 to 2017 \\
			Age at diagnosis & Numerical & 0 to 88 \\
			Race & Categorical & American Indian/Alaska Native, Asian or Pacific Islander, Black, White, Unknown \\
			ICD-10 Code & Categorical & Main bronchus, Upper lobe, Middle lobe, Lower lobe, Overlapping lesion, Lung (NOS) \\
			
			\multirow[t]{4}{*}{Histology} & \multirow[t]{4}{*}{Categorical}& Adenocarcinoma, Bronchiolo-alveolar carcinoma, non-mucinou \\
			& & Invasive mucinous adenocarcinoma, Adenocarcinoma with mixed subtypes \\
			& & Papillary adenocarcinoma (NOS), Clear cell adenocarcinoma (NOS) \\
			& & Mucinous adenocarcinoma, Signet ring cell carcinoma, Acinar cell carcinoma \\
			Laterality & Categorical & Left, Right, Other \\
			\hline			
			\bottomrule
		\end{tabular}
	\end{adjustbox}
	\caption{A list of features common between the GDC and SEER datasets with their data types and valid value ranges.}
	\label{table:common_columns}
\end{table}

\section{Discussion of Results}
\label{sec:discussion}

\subsection{MNIST}
\label{sec:results:mnist}
The results in Table \ref{table:mnist} show the $F_1$ classification performance for the images with missing data and with both basic AE and VAE augmentation. Twenty six total experiments were performed across thirteen common columns combinations on both sides of the images with one of the augmentation methods providing twenty of the best results. The basic AE usually did better with fewer common columns compared with the VAE which may suggest that the more complicated VAE network benefits more with additional data. While the overall improvements of augmentation do not seem to be very significant, they still represent a large portion of the remaining performance as most of the $F_1$ scores are already above 99 percent.

\begin{table}[h!]
	\centering
	\renewcommand{\arraystretch}{1.25}
	\begin{adjustbox}{center,max width = 200mm}
		\footnotesize
		\begin{tabular}{*{8}{c}}\bottomrule
			\multicolumn{8}{c}{MNIST Dataset}\\
			\bottomrule
			Common & \multirow{2}{*}{A Only} & \multirow{2}{*}{A + B* AE} & \multirow{2}{*}{A + B* VAE} & & \multirow{2}{*}{B Only} & \multirow{2}{*}{A* + B AE} & \multirow{2}{*}{A* + B VAE} \\
			Columns & & & & & & & \\
			\hline
			2 & 97.16 & \textbf{97.17} & 97.08 & & \textbf{96.50} & 96.42 & 96.30 \\
			4 & \textbf{97.70} & 97.66 & 97.62 & & \textbf{97.48} & 97.42 & 97.31 \\
			6 & 98.02 & \textbf{98.12} & 98.04 & & 98.10 & \textbf{98.18} & 98.04 \\
			8 & 98.42 & \textbf{98.46} & 98.36 & & 98.57 & \textbf{98.73} & 98.58 \\
			10 & 98.61 & \textbf{98.70} & 98.64 & & 98.87 & \textbf{98.93} & 98.92 \\
			12 & 98.86 & 98.86 & \textbf{98.93} & & 99.07 & \textbf{99.11} & 98.99 \\
			14 & 98.96 & \textbf{99.02} & 98.97 & & 99.09 & \textbf{99.14} & 99.08 \\
			16 & \textbf{99.11} & 99.08 & 99.09 & & 99.15 & 99.15 & \textbf{99.17} \\
			18 & \textbf{99.11} & 99.08 & 99.09 & & 99.11 & 99.14 & \textbf{99.17}\\
			20 & 99.10 & 99.05 & \textbf{99.15} & & 99.11 & 99.14 & \textbf{99.20} \\
			22 & \textbf{99.11} & 99.03 & 99.10 & & 99.13 & 99.15 & \textbf{99.16} \\
			24 & 99.07 & 99.05 & \textbf{99.13} & & 99.15 & 99.13 & \textbf{99.17} \\
			26 & 99.08 & \textbf{99.10} & 99.08 & & 99.09 & 99.10 & \textbf{99.16} \\
			\bottomrule
		\end{tabular}
	\end{adjustbox}
	\caption{The $F_1$ scores for the MNIST classifier experiments with the left and right hand data as the primary datasets. The $*$ symbol marks the synthetic features that were generated with either the AE or VAE based networks.}
	\label{table:mnist}
\end{table}

\begin{figure}[h] 
	\centering
	\includegraphics[width=\textwidth]{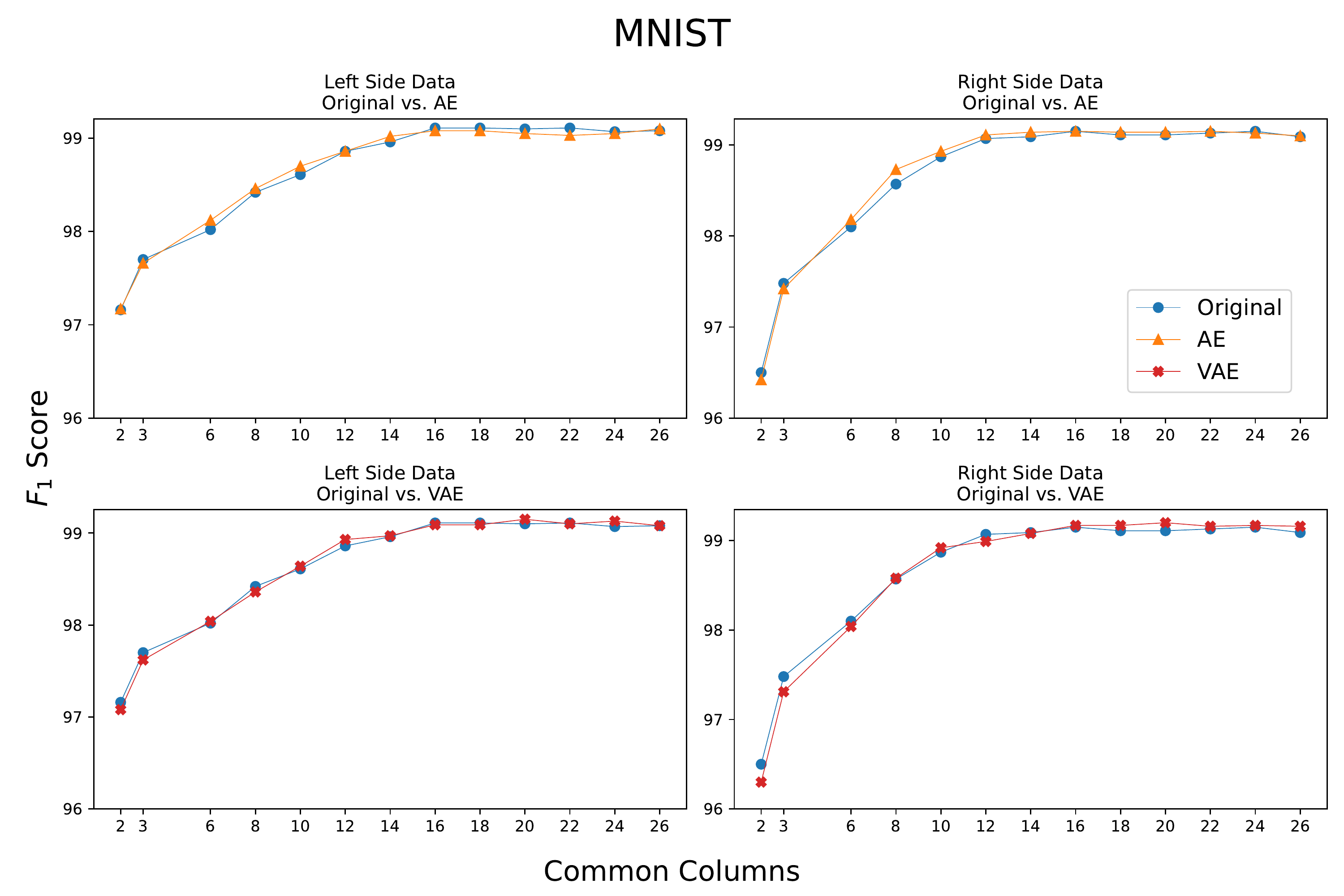}
	\caption{Plot showing the comparison of MNIST images with missing columns against the AE and VAE based augmentation methods.}
	\label{fig:mnist_plot}
\end{figure}

Figure \ref{fig:mnist_plot} shows line graph of the performance of the MNIST images with missing columns against the AE and VAE based augmentation methods. As expected, including more common columns makes the problem easier for both the missing data case and with augmentation but performance flattens out after enough data is available. Because of how the handwritten digit were presented, there is almost no useful information at the far left and right sides of the original images and most of the information is in the middle. The first few common columns are in this critical section which explains why adding just a few columns makes a significant impact on the result. This is most apparent with the AE plots where most of the gains were found within the first eight common columns.

\begin{figure}
	\centering
	\includegraphics[width=\linewidth,trim= 5 5 5 5,clip]{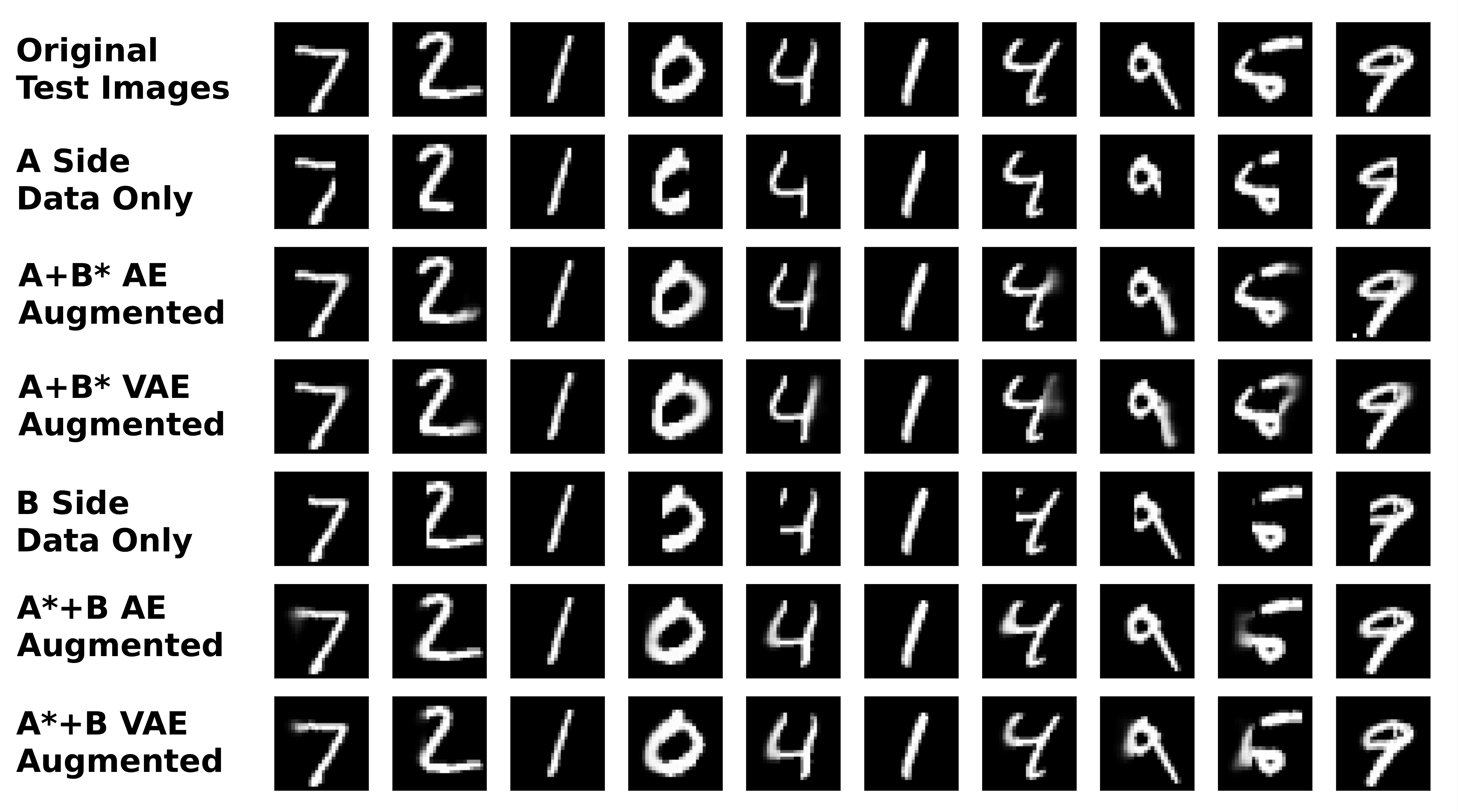}
	\caption{Comparison of augmented images for both the A and B side of the MNIST dataset. In the augmentation results, A* and B* refers to their corresponding synthetic representations. These images were generated using eight common columns.}
	\label{fig:mnist_images}
\end{figure}

Figure \ref{fig:mnist_images} show examples of the generated digits for the \textit{A} and \textit{B} sides of the images. In most of these cases, the AE and VAE algorithms did a reasonably good job replacing the missing image columns. The largest discrepancy between these methods was with the second to last image column for the number five. This was a more difficult problem as the digit was poorly written and the bottom semicircle was completely closed. The AE got much closer when attempting to complete the right hand size of the image as the VAE may have mistook the image as a partially completed eight. A similar problem occurred when completing the left hand side as it could be seen as a six. While the VAE did appear to make some of the clearest augmentations, its mistakes were more pronounced but might be improved with larger training sets.

\subsection{CIFAR10}
\label{sec:results:cifar10}
Like with MNIST, Table \ref{table:cifar10} shows that the data augmentation process improved $F_1$ scores for 23 of the 30 cases across the AE and VAE experiments. The impact of augmentation was more pronounced with this dataset which may be attributed to the increased complexity of the data. Many of the experiments showed improvements in the range of 0.5 to 1.0\%. The AE method gave most of the top results, although the VAE outperformed the non-augmented data in the majority of experiments. Augmentation tended to work better when there was enough data to learn from and some critical information was still missing that needed to be replaced. This ideal range was around 8 - 24 common columns for CIFAR10.

\begin{table}[h!]
	\centering
	\renewcommand{\arraystretch}{1.25}
	\begin{adjustbox}{center,max width = 200mm}
		\footnotesize
		\begin{tabular}{*{8}{c}}\bottomrule
			\multicolumn{8}{c}{CIFAR10 Dataset}\\
			\bottomrule
			Common & \multirow{2}{*}{A Only} & \multirow{2}{*}{A + B* AE} & \multirow{2}{*}{A + B* VAE} & & \multirow{2}{*}{B Only} & \multirow{2}{*}{A* + B AE} & \multirow{2}{*}{A* + B VAE} \\
			Columns & & & & & & & \\
			\hline
			2 & \textbf{64.50} & 64.38 & 64.13 &  & \textbf{64.15} & 64.07 & 63.67 \\
			4 & 65.95 & 65.52 & \textbf{65.96} &  & \textbf{65.37} & 65.15 & 64.83 \\
			6 & 66.14 & 66.01 & \textbf{66.33} &  & 65.30 & \textbf{65.83} & 65.70 \\
			8 & 66.80 & 66.99 & \textbf{67.48} &  & 66.02 & \textbf{66.75} & 66.43 \\
			10 & 67.32 & 67.01 & \textbf{67.80} &  & 66.47 & \textbf{66.89} & 66.68 \\
			12 & 67.68 & \textbf{68.07} & 67.54 &  & 67.19 & \textbf{67.51} & 67.36 \\
			14 & 68.07 & \textbf{68.74} & 68.14 &  & 67.23 & \textbf{67.82} & 67.69 \\
			16 & \textbf{68.94} & 68.88 & 68.62 &  & 67.98 & 68.46 & \textbf{68.80} \\
			18 & 68.58 & 68.95 & \textbf{69.10} &  & \textbf{68.63} & 68.60 & 68.31 \\
			20 & 68.69 & \textbf{69.26} & 69.13 &  & 68.59 & \textbf{69.02} & 68.58 \\
			22 & 69.29 & \textbf{69.50} & 69.18 &  & 68.96 & \textbf{69.44} & 69.04 \\
			24 & 69.39 & \textbf{70.03} & 69.92 &  & 69.09 & \textbf{70.11} & 69.38 \\
			26 & 69.80 & 69.49 & \textbf{70.38} &  & 69.66 & 69.51 & \textbf{69.77} \\
			28 & 69.81 & \textbf{70.14} & 70.10 &  & \textbf{69.99} & 69.66 & 69.92 \\
			30 & 70.12 & 70.32 & \textbf{70.38} &  & \textbf{69.90} & 69.77 & 69.79 \\						
			\bottomrule
		\end{tabular}
	\end{adjustbox}
	\caption{The $F_1$ scores for the CIFAR10 classifier experiments. The $*$ symbol marks the synthetic features that were generated with either the AE or VAE based networks.}
	\label{table:cifar10}
\end{table}

\begin{figure}[h] 
	\centering
	\includegraphics[width=\textwidth]{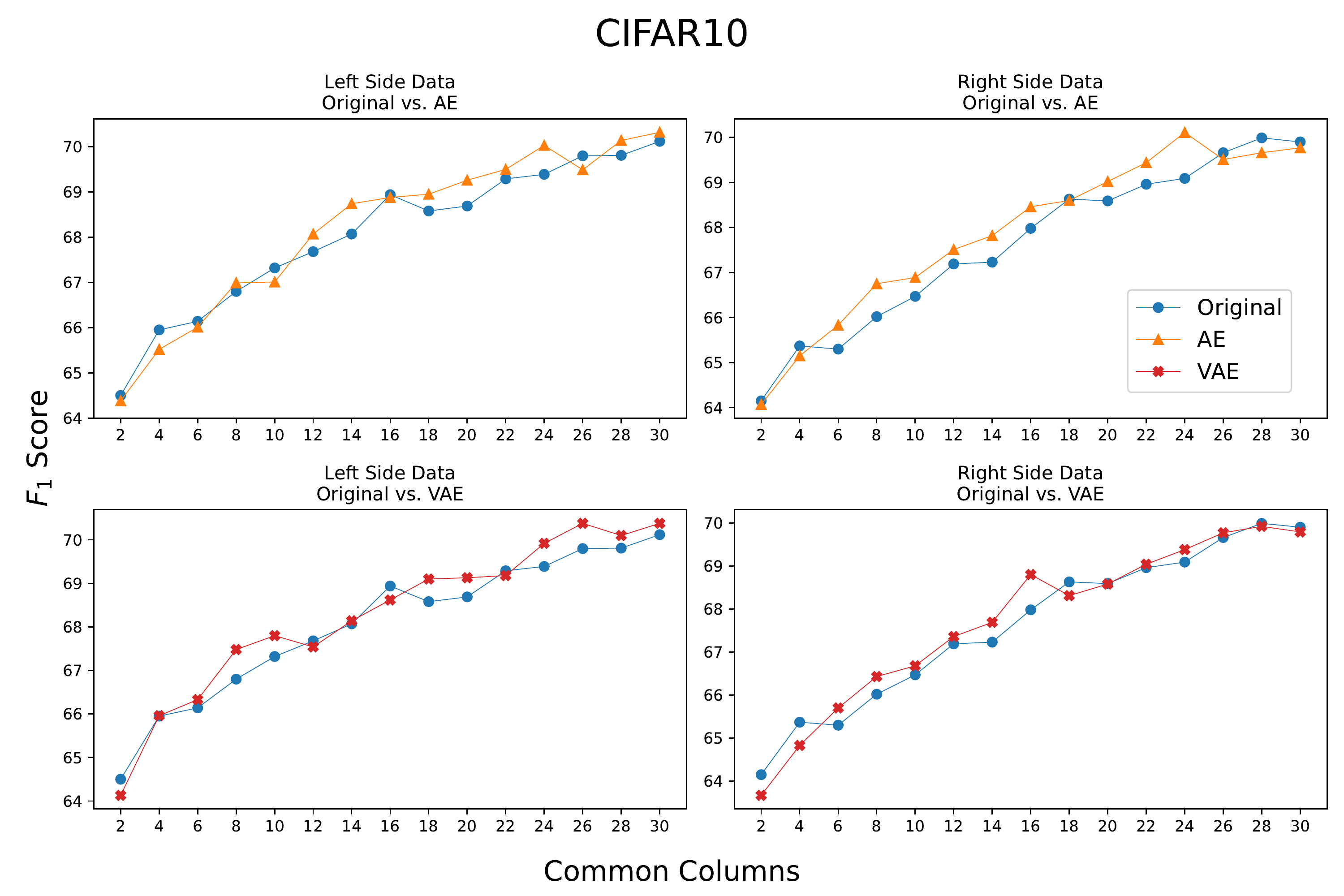}
	\caption{Plot showing the comparison of CIFAR10 images with missing columns against the AE and VAE based augmentation methods.}
	\label{fig:cifar10_plot}
\end{figure}

The value of data augmentation for CIFAR10 is more evident as shown in Figure \ref{fig:cifar10_plot}. $F_1$ scores increase when more data is included but does not plateau like with MNIST as there is useful data throughout the images. Both augmentation methods improve scores in most cases and the relative performance increase is present as more columns are added.

\begin{figure}
	\centering
	\includegraphics[width=\linewidth,trim= 5 5 5 5,clip]{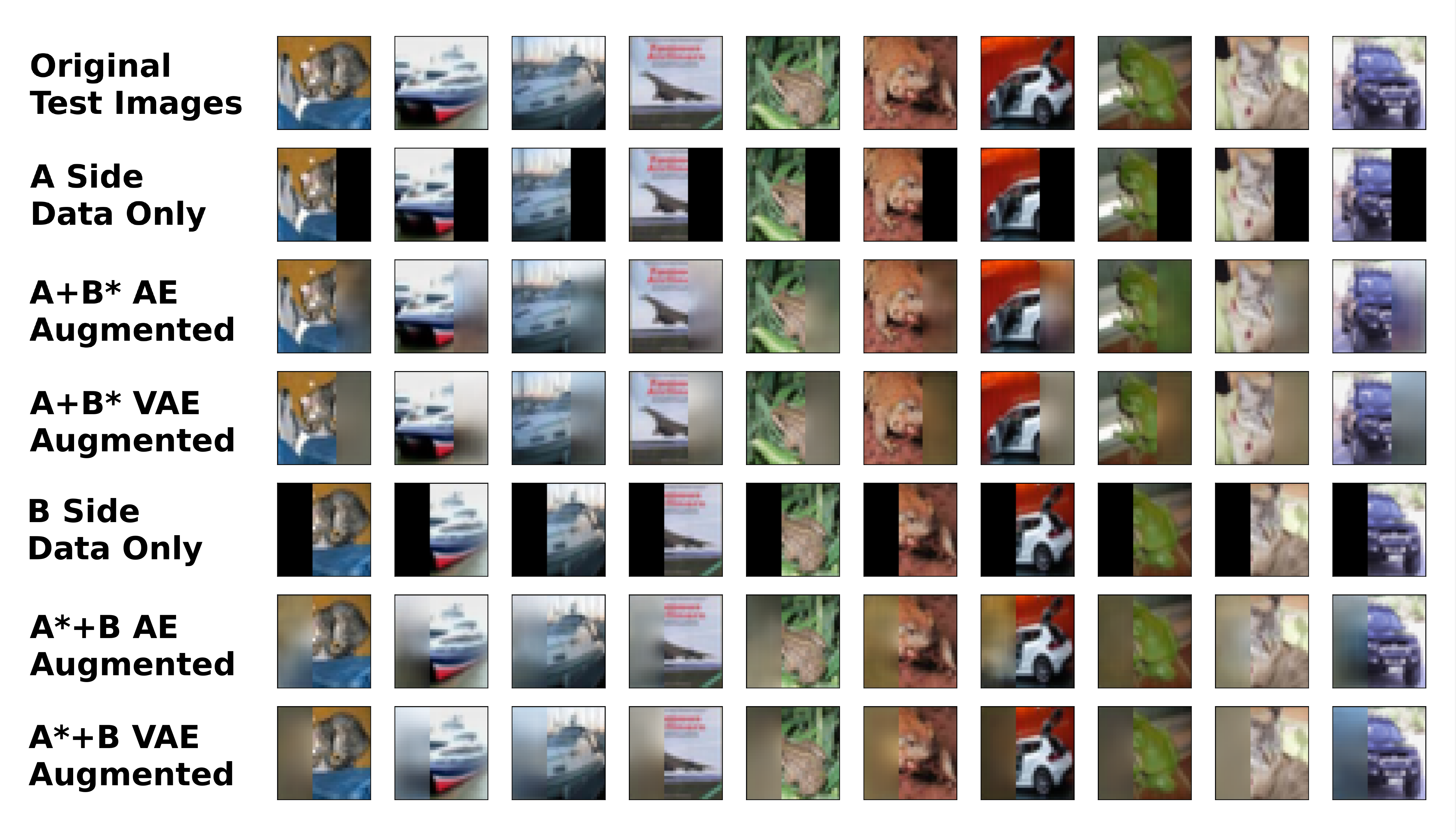}
	\caption{Comparison of augmented images for both the A and B side of the CIFAR10 dataset. In the augmentation results, A* and B* refers to their corresponding synthetic representations. These images were generated using eight common columns.}
	\label{fig:cifar10_images}
\end{figure}

Figure \ref{fig:cifar10_images} show examples of the generated digits for the \textit{A} and \textit{B} sides of the images. Although augmentation improved $F_1$ scores, the quality of the generated image data was much worse than with MNIST. The CIFAR10 dataset is not very large compared to recent deep learning image datasets and with its size further reduced to create the non-overlapping \textit{A} and \textit{B} datasets. While there may not be enough training data to produce high quality image data, it was still enough to help the classifier. In some cases, such as the boats depicted in columns two and tree of Figure \ref{fig:cifar10_images}, the new image data does look like a very blurry version of the true image data. However, the frog images in columns 5, 6, and 8 provide almost no information but may be attributed to the diverse coloring, backgrounds, or orientations. Boats, on the other hand, tend to look more similar as they often share common colors and backgrounds which may require fewer training examples.

\subsection{GDC and SEER}
\label{sec:gdc-seer-results}
As mentioned in Section \ref{sec:experiments:nihseer}, these experiments used 522 and 647k examples for the GDC and SEER datasets respectively. Unlike the prior image based experiments, the number of common columns were naturally presented as the two datasets were already separated. Because the GDC dataset was much smaller, the SEER dataset was randomly undersampled to evaluate the impact that the dataset size has on the augmentation process. Classification was performed to predict if a patient would survive 24 months or longer after lung cancer treatment.

Table \ref{table:nih_seer} shows the $F_1$ scores for the experiments performed with different level of SEER undersampling. Although using a high percentage of the SEER dataset did not help the GDC classification, using 500 to 10k SEER examples did. This resulted in both datasets being more balanced and using much more SEER examples in the training process could make the latent space used for decoding too cluttered to generate useful features. When using the 10k SEER examples for training, the AE network increased $F_1$ score by almost 2.3\%.

When augmenting the SEER dataset, the VAE  outperformed the AE method in all cases and improved the overall $F_1$ scores six out of eight times compared to just using the original data. As expected, including more of the original SEER data improved the non-augmented classification results. However, augmentation started to consistently improve results when at least 10k SEER examples were used.

These results show that cross-database feature augmentation can improve classifier performance, especially when the larger dataset is the one being augmented. One potential benefit from this kind of augmentation is that good results may be achievable without the entire dataset. This can be more important for cases where datasets are very large and are slow to train or when it is difficult to acquire large training dataset, which is a common challenge with medical data. Augmenting only 100k SEER examples with the small GDC dataset provided better results than the entire SEER dataset without augmentation.

The smaller GDC dataset got less benefit from augmentation but did see some improvement with a smaller selection of SEER examples. Additional tuning of the autoencoding network and feature engineering may result in a more significant impact with even smaller datasets.

\begin{table}[h!]
	\centering
	\renewcommand{\arraystretch}{1.25}
	\begin{adjustbox}{center,max width = 100mm}
		\footnotesize
		\begin{tabular}{lccccccc}\bottomrule
			\multicolumn{8}{c}{GDC and SEER Tabular Datasets}\\
			\bottomrule
			SEER Count & GDC & GDC + SEER* AE & GDC + SEER* VAE & & SEER & GDC* + SEER AE & GDC* + SEER VAE \\
			\hline
			250 & \bf{83.74} & 82.16 & 79.36 & & \bf{77.22} & 69.63 & 75.61 \\
			500 & 79.25 & \bf{79.49} & 77.76 & & 77.09 & 75.00 & \bf{77.14} \\
			1k & 82.62 & \bf{83.00} & 79.61 & & \bf{77.36} & 73.41 & 75.82 \\
			10k	& 84.61	& \bf{86.88} & 74.63 & & 78.94 & 77.71 & \bf{79.87} \\
			50k & 89.29 & \bf{89.58} & 82.45 & & 81.83 & 81.67 & \bf{82.96} \\
			100k & \bf{81.96} & 81.56 & 77.18 & & 82.73 & 82.23 & \bf{83.93} \\
			250k & \bf{86.01} & 84.36 & 83.38 & & 82.21 & 82.00 & \bf{83.61} \\
			500k & \bf{84.91} & 83.48 & 81.23 & & 83.10 & 82.92 & \bf{84.26} \\
			647k & \bf{84.83} & 80.33 & 83.27 & & 83.42 & 83.11 & \bf{84.60} \\			
			\bottomrule
		\end{tabular}
	\end{adjustbox}
	\caption{$F_1$ scores for the GDC and SEER experiments. Since the SEER dataset is much larger than GDC, experiments were run with different subsets with the total included examples in the \textit{SEER Count} column. These two datasets were augmented in the same manner for MNIST and CIFAR10.}
	\label{table:nih_seer}
\end{table}

\begin{figure}[h] 
	\centering
	\includegraphics[width=1\textwidth]{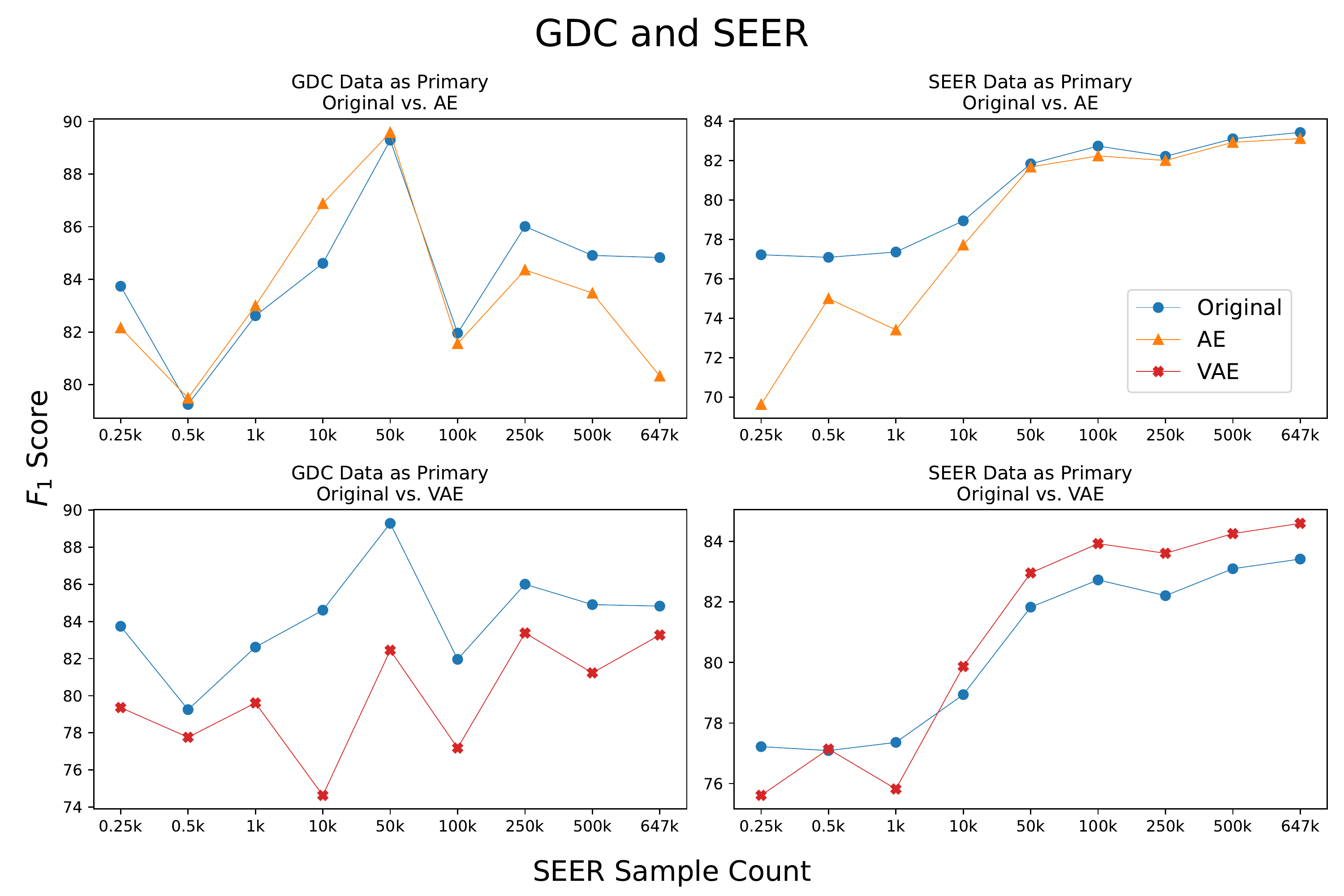}
	\caption{Plot showing the comparison of augmentation with the GDC and SEER datasets using the AE and VAE networks.}
	\label{fig:gdc_seer_plot}
\end{figure}

\subsection{Future Work}
In addition to the work presented in this paper, there are a number of other topics that should be further investigated. The data cleaning process performed for the GDC and SEER datasets removed any examples that had many missing features and entire features that had a significant number of missing values to ensure the cleanest possible data. However, some of this removed data could be used if a more permissive imputation process was used such as the traditional statistical based replacement or even within the autoencoder process itself.

Class imbalance can negatively impact classifier performance as it can be biased towards the majority class which can also affect deep learning problems \cite{pulgar2017impact, buda2018systematic, johnson2019survey}. Although the MNIST and CIFAR10 datasets were mostly balanced, the GDC and SEER datasets has a much higher imbalance with the two year survival examples representing approximately 72\% of the dataset. The proposed feature generation system could also be extended to create entirely new features for class balancing as previously done with a VAE based network \cite{wan2017variational}. The potential benefit of majority class undersampling, minority class oversampling, or some combination of both could be further pursued using both traditional machine learning and deep learning algorithms. 

Instance level difficulty is another approach used to improve model performance but was not considered in this work. The data augmentation method may benefit by focusing more on specific types of examples. Most of the current work on instance level difficulty has focused on a single dataset so there is not much research on how the difficulty of examples between multiple datasets would affect mutual encodings.

In this work, only one dataset was used to augment another but there may be cases where the augmentation process could benefit from multiple datasets. Each extra dataset could be used to generate different groups of features to augment the target dataset. There could also be features common between datasets \textit{A} and \textit{B} but different features common between \textit{B} and \textit{C}. 

The simpler AE and VAE networks used in the experiments could be replaced with deeper networks of the same type or more complicated architectures like stacked autoencoders and GANs. These neural networks often perform better on datasets with many more examples or features such as high resolution pictures, 3D medical images, or gene expression data.  In addition to classification, there are other types of machine learning algorithms that could benefit from the proposed technique such as regression and clustering.

\section{Conclusion}
\label{sec:conclusion}
Within many problem domains, there are often many related datasets that do not have matching features making the curation of large datasets difficult. In this work, we have proposed an autoencoder based solution for data augmentation using common features between these disparate datasets. As long as these datasets are contextually similar, the information unique from each dataset could be used to improve the performance of classifiers. We have shown that this approach can work with both image and tabular data as well as different types of autoencoder networks.

Although our experiments used relatively shallow autoencoders, they could be replaced with much complicated or problem specific architectures such as GANs, stacked autoencoders, and deep pre-trained models. In addition to the MNIST and CIFAR10 examples, we showed that the data augmentation can work on real world medical datasets. Even the small GDC dataset was able to provide a benefit for classifying the much larger SEER dataset. This method is especially useful for domains like medicine which often is limited to small datasets collected as part of individual studies.

This initial work on data augmentation using common features has suggested several new areas of research. This approach could be extended to include the use of multiple datasets for augmentation, larger autoencoder style networks, and different styles of imputation. Utilizing data properties like class imbalance or instance level difficulty may provide further benefits.





\bibliographystyle{elsarticle-num}
\bibliography{references}   
\end{document}